\documentclass[a4paper,twoside]{article}

\usepackage{epsfig}
\usepackage{subcaption}
\usepackage{calc}
\usepackage{amssymb}
\usepackage{amstext}
\usepackage{amsmath}
\usepackage{amsthm}
\usepackage{multicol}
\usepackage{pslatex}
\usepackage{apalike}
\usepackage{microtype}                 
\usepackage{url}
\usepackage[inline]{enumitem}
\usepackage{gensymb}
\usepackage[hidelinks]{hyperref}
\usepackage{fancyhdr}
\usepackage{SCITEPRESS}     

\graphicspath{{figures/}{pictures/}{images/}{./}} 
\setlist[enumerate]{itemsep=0mm}

\begin{document}

\title{Combining Local and Global Pose Estimation for Precise Tracking of Similar Objects}

\author{\authorname{Niklas Gard\sup{1}\orcidAuthor{0000-0002-0227-2857}, Anna Hilsmann\sup{1}\orcidAuthor{0000-0002-2086-0951} and Peter Eisert\sup{1,2}\orcidAuthor{0000-0001-8378-4805}}  
\affiliation{\sup{1}Vision and Imaging Technologies, Fraunhofer HHI, Einsteinufer 37, 10587 Berlin, Germany}
\affiliation{\sup{2}Institute for Computer Science, Humboldt University of Berlin, Unter den Linden 6, 10099 Berlin, Germany}
\email{\{ niklas.gard , anna.hilsmann, peter.eisert \}@hhi.fraunhofer.de}
}


\abstract{
In this paper, we present a multi-object 6D detection and tracking pipeline for potentially similar and non-textured objects. The combination of a convolutional neural network for object classification and rough pose estimation with a local pose refinement and an automatic mismatch detection enables direct application in real-time AR scenarios. A new network architecture, trained solely with synthetic images, allows simultaneous pose estimation of multiple objects with reduced GPU memory consumption and enhanced performance.
In addition, the pose estimates are further improved by a local edge-based refinement step that explicitly exploits known object geometry information. For continuous movements, the sole use of local refinement reduces pose mismatches due to geometric ambiguities or occlusions.
We showcase the entire tracking pipeline and demonstrate the benefits of the combined approach. Experiments on a challenging set of non-textured similar objects demonstrate the enhanced quality compared to the baseline method. 
Finally, we illustrate how the system can be used in a real AR assistance application within the field of construction. 
}

\onecolumn \maketitle \normalsize \setcounter{footnote}{0} \vfill
\vfill
\thispagestyle{fancy}

\section{\uppercase{Introduction}}

\begin{figure*}
\centering  
\begin{subfigure}{.14\linewidth }
  \centering
  \includegraphics[width=\linewidth]{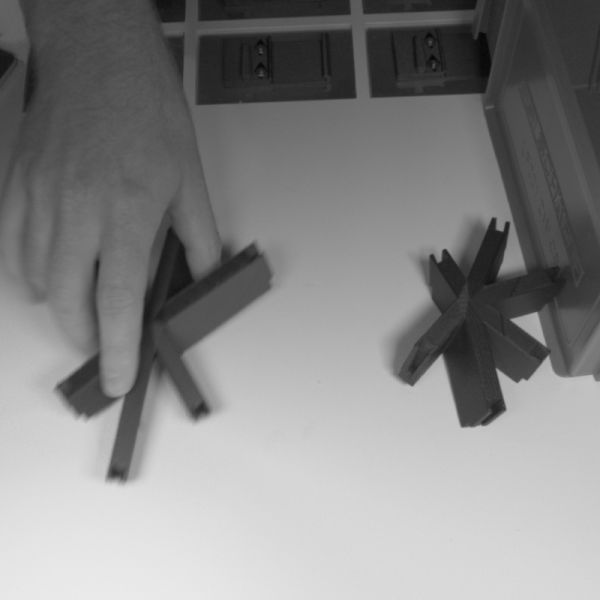}
\end{subfigure}
\hspace{.01\linewidth}
\begin{subfigure}{.14\linewidth }
  \centering
  \includegraphics[width=\linewidth]{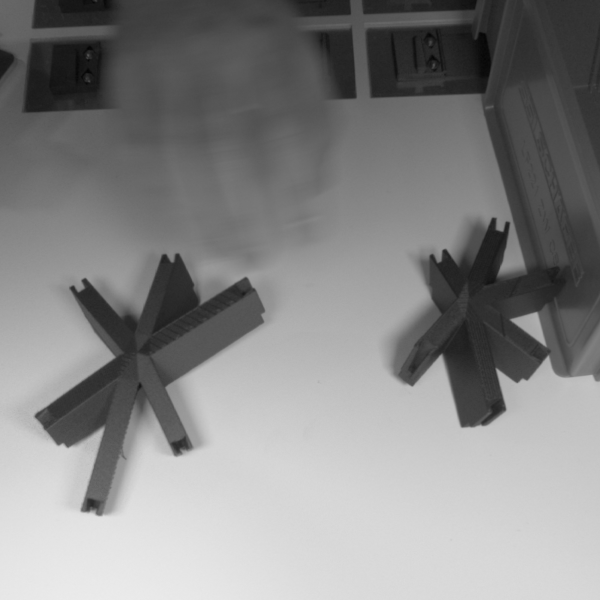}
\end{subfigure}
\hspace{.01\linewidth}
\begin{subfigure}{.14\linewidth }
  \centering
  \includegraphics[width=\linewidth]{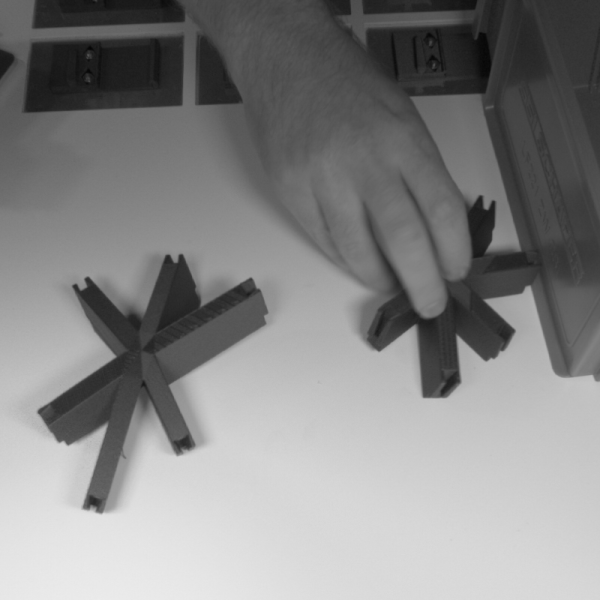}
\end{subfigure}
\hspace{.01\linewidth}
\begin{subfigure}{.14\linewidth }
  \centering
  \includegraphics[width=\linewidth]{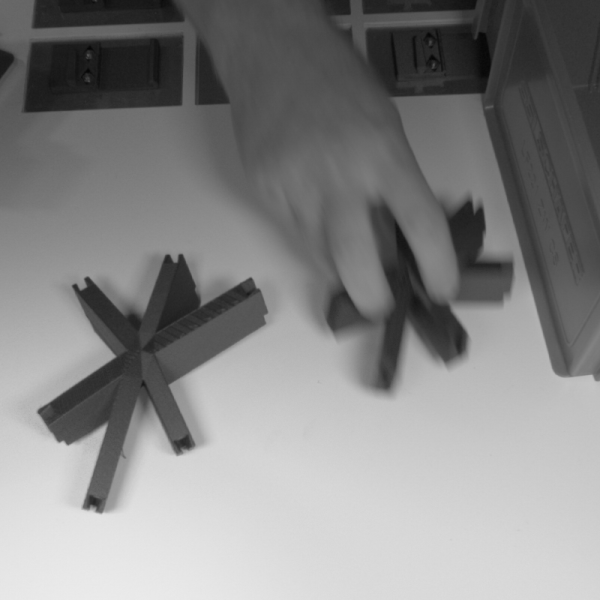}
\end{subfigure} \\

\begin{subfigure}{.14\linewidth }
  \centering
  \includegraphics[width=\linewidth]{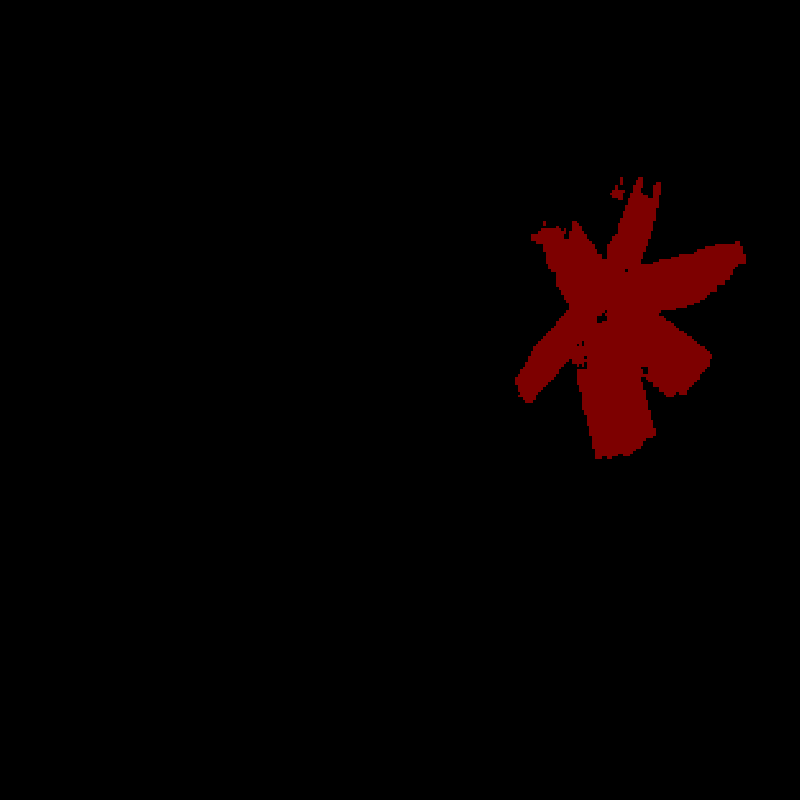}
\end{subfigure}
\hspace{.01\linewidth}
\begin{subfigure}{.14\linewidth }
  \centering
  \includegraphics[width=\linewidth]{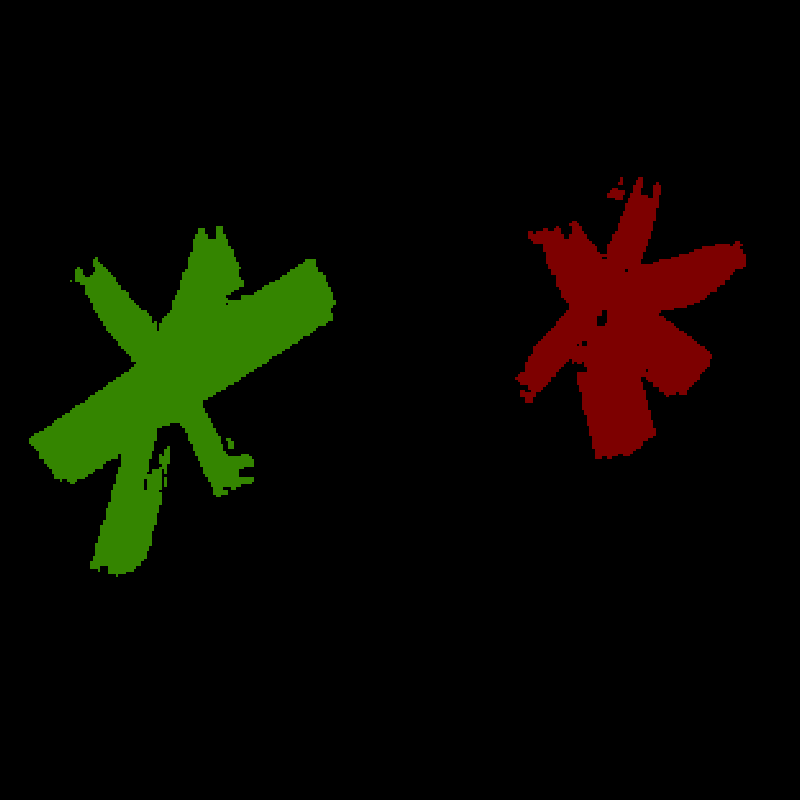}
\end{subfigure}
\hspace{.01\linewidth}
\begin{subfigure}{.14\linewidth }
  \centering
  \includegraphics[width=\linewidth]{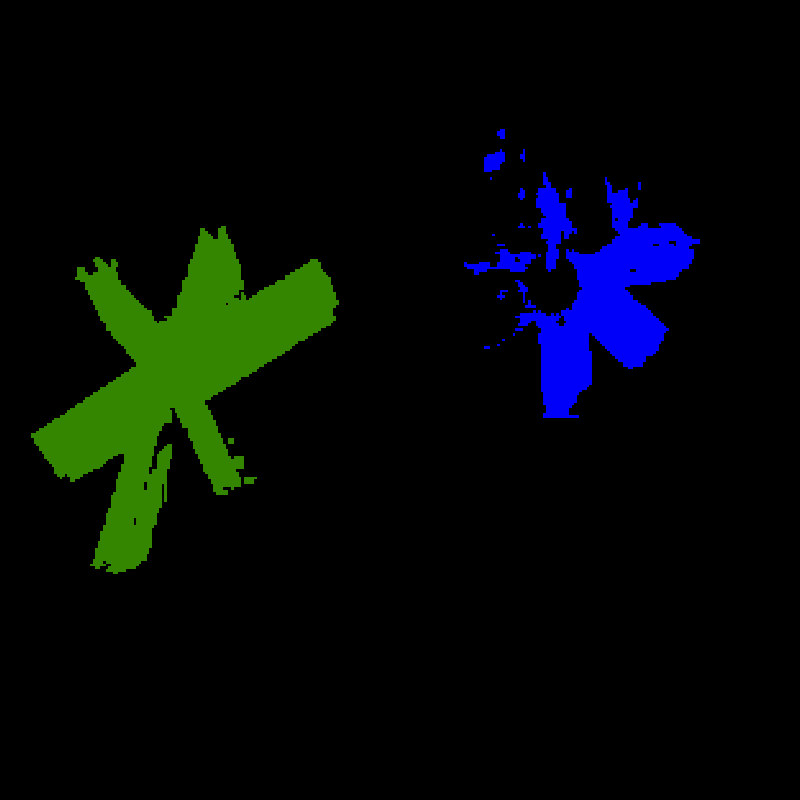}
\end{subfigure}
\hspace{.01\linewidth}
\begin{subfigure}{.14\linewidth }
  \centering
  \includegraphics[width=\linewidth]{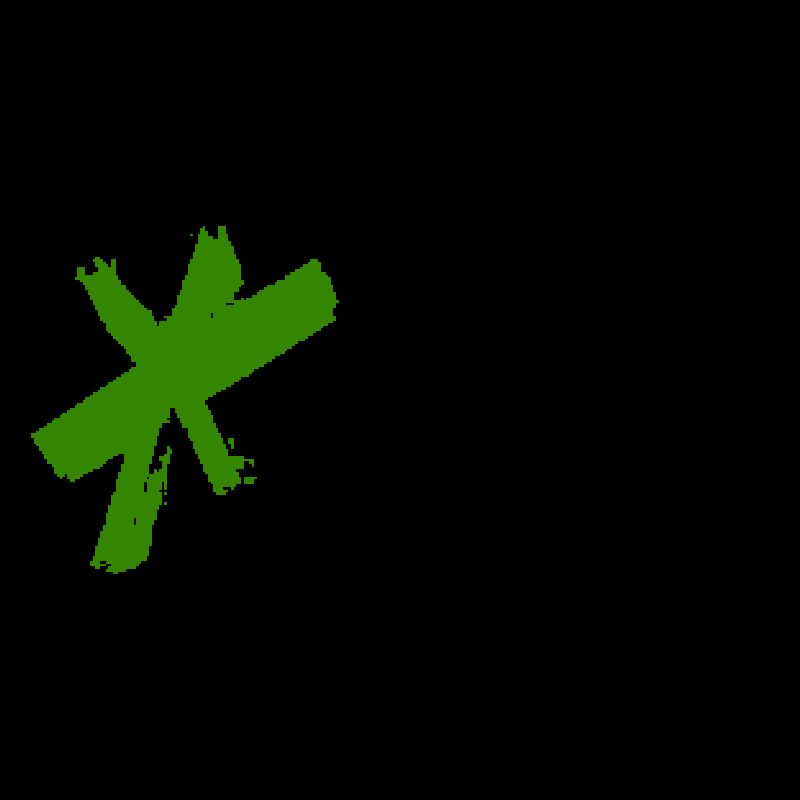}
\end{subfigure} \\
\begin{subfigure}{.14\linewidth }
  \centering
  \includegraphics[width=\linewidth]{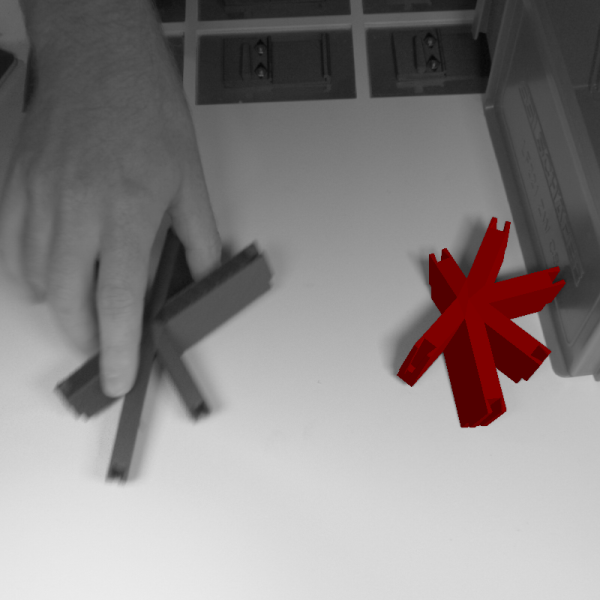}
    \caption{}
       \label{fig:tracking1}
\end{subfigure}
\hspace{.01\linewidth}
\begin{subfigure}{.14\linewidth }
  \centering
  \includegraphics[width=\linewidth]{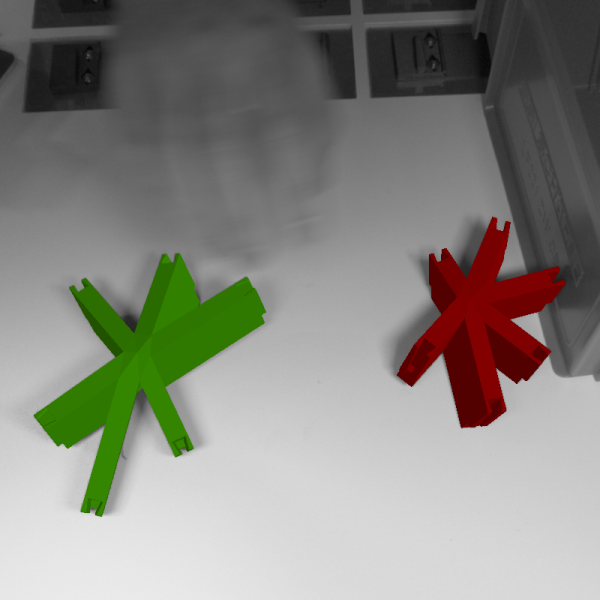}
      \caption{}
  \label{fig:tracking2}
\end{subfigure}
\hspace{.01\linewidth}
\begin{subfigure}{.14\linewidth }
  \centering
  \includegraphics[width=\linewidth]{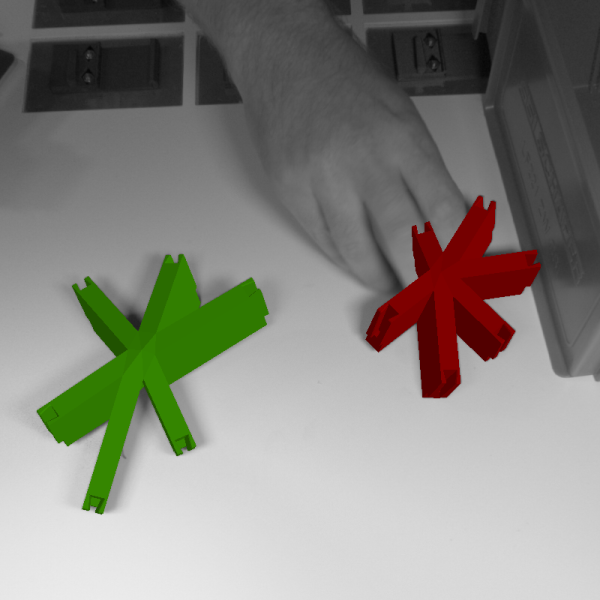}
      \caption{}
   \label{fig:tracking3}
\end{subfigure}
\hspace{.01\linewidth}
\begin{subfigure}{.14\linewidth }
  \centering
  \includegraphics[width=\linewidth]{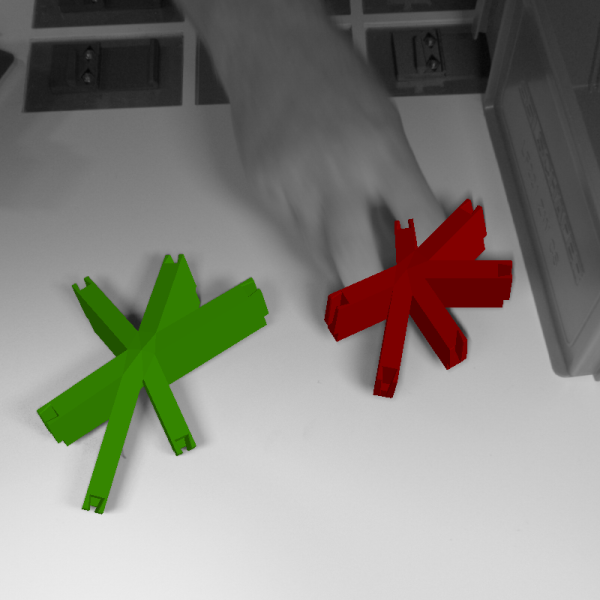}
   \caption{}
   \label{fig:tracking4}

\end{subfigure}
\centering  
\caption{After the left object is uncovered (a, b) tracking starts immediately. Local tracking keeps active for the right object also if the CNN detects a wrong (c) or no object (d). Rows from top to bottom: camera image, estimated semantic segmentation, camera image with rendered overlay.}
\label{fig:tracking}
\end{figure*}

The detection and subsequent registration of 3D rigid objects in videos are key components in augmented reality (AR) systems. When interacting with real-world objects, their relative pose with respect to a camera has to be known to enrich them with additional information. 
Specifically, there are two subtasks: global 6D pose estimation without prior knowledge of the pose and local frame-to-frame tracking where an initial pose is known from the previous frame. This paper connects current literature on the research areas in a dynamic system and contributes new ideas to both to balance them for optimal AR usage. 

We target challenging industrial or construction scenarios, involving manual object assembly or sorting. 
They require stable detection and tracking, as well as real-time capability and occlusion handling.
In addition, the objects are often untextured, similar in shape and color, and produced in small batches. 

This presents particular hurdles for pose estimation. While convolutional neural network(CNN)-based 6D pose estimators from single-camera images have advanced rapidly in recent times, their ability to perform multi-object pose estimation has received only limited attention. Often best results for multiple objects are achieved by training a separate network for each object \cite{Song2020,Park2019,Peng2019}. Not only does this require more memory than a single-network solution, but also classification is more difficult. Either each network must be tested if an object of unknown type is visible, or even another network must be trained to identify the objects. However, knowledge of multiple similar objects helps distinguish them and prevents false-positive estimates. We extend a well-known pose estimator PVNet \cite{Peng2019} by using object-specific parameters locally within the estimated semantic masks. This improves the distinguishing of similar objects with a single network and the handling of shape ambiguities.

Next, objects produced in small batches make the collection of real-world training images infeasible. The use of synthetic renderings allows to create unlimited amounts of perfectly labeled data to flexibly add new objects to the system. We see local refinement as a key to solving different problems. First, the so-called domain gap limits the accuracy of CNNs trained on synthetic data tested on real data \cite{Wang2020}.
Second, a CNN usually processes images with limited resolution, limiting also the accuracy of the pose estimates. Contrary to the trend to make the training data more realistic \cite{Hodan2019}, refinement narrows the domain gap and enables us to use simple domain randomized training data \cite{To2018}. Even only single object images are used for training, although the use in a multi-object system is intended. The local edge-based refinement matches the 3D model with image edges on high-resolution images.
Third, false-positive detections may arise from the CNN due to shape ambiguities.
Object edges provide a cue to suppress them. Poses are validated only if it is possible to accurately match the projected contour of the 3D model with the image edges, using an edge deviation error from local pose refinement.

Even though the CNN-based solutions are real-time capable, we show that it is often beneficial to prioritize local refinement in video sequences (\autoref{fig:tracking3},\subref{fig:tracking4}), e.g.\ in situations not explicitly modelled in the training data. 
Only if local refinement fails due to occlusion or fast motion, edge-based pose validation actively triggers reinitialization (\autoref{fig:tracking1},\subref{fig:tracking2}), so that otherwise no CNN evaluation is required.

In summary, we present a pipeline for tracking and 6D detection of multiple similar objects in AR systems, with automatable training, and evaluate the individual components with 13 similar objects from a real AR application. We demonstrate our pipeline on synthetic images with domain gap and real video sequences and show benefits and limitations of individual components and possibilities of the overall system.

\section{\uppercase{Related Work}}
\label{sec:related work}

\paragraph{Augmented reality for assembly.}
Providing guidance e.g.\ via head-mounted displays during assembly and construction tasks is an essential AR use case, motivated by the reduced time needed to complete a task compared to paper manuals \cite{Henderson2010}. User acceptance is an important challenge \cite{Masood2020}, and reliable tracking is a key factor to achieving immersive, easy-to-use systems. While stable localization is already integrated into widely used hardware \cite{Vassallo2017},  pipelines with easy automatic extendibility for new objects as well as stable 6D tracking for AR are still rare.  \cite{Zubizarreta2019} provide a framework based on chamfer matching with conic priors to detect and register CAD models of machine-made objects in monocular images. A limitation is that objects have to consist of conics to be recognized, and it is unclear whether the system works for geometrically similar objects.
Recognition of subtle differences is studied in the field of assembly state detection \cite{Liu2020}.
Assembly state detection has also been paired with pose estimation \cite{Su2019}, but without combining it with frame-to-frame pose refinement, it does not provide stable tracking for AR applications. 

\paragraph{Multi-object 6D pose estimation.}
A common way to infer the pose of known objects from a monocular image is to use a CNN to find the location of the 2D image projection of 3D points, e.g.\ object-specific keypoints or dense coordinate maps, and estimate the pose with a Perspective-n-Point (PnP) algorithm \cite{Peng2019,Song2020,Zakharov2020,Tremblay2018,Zhigang2019,Park2019}. 

As an example, PVNet \cite{Peng2019} segments objects and simultaneously predicts unit vector fields inside the estimated mask pointing towards the 2D projections of keypoints.
The intersection of two randomly selected points leads to a 2D estimate, which is validated with a RANSAC-based voting scheme. This generates a high level of robustness against occlusions. To detect multiple objects with a single network, they propose to simply increase the number of classes and also the number of estimated vector fields.
Nevertheless, with PVNet and the other approaches mentioned above, the best results are achieved when a single network is trained for every object. In addition, the amount of GPU memory needed in training drastically increases with the number of outputs, and training becomes slower and more difficult.
 
\cite{Sock2020} describe a performance drop due to scalability problems for a backbone that uses a similar trivial multi-object extension \cite{Rad2017}. To close this gap, they add object-specific normalization parameters to the CNN using Conditional Instance Normalization (CIN) \cite{Dumoulin2017}. With CIN, the right normalization parameters can only be selected if the object identity information is known, e.g.\ by using a bounding box detector, and only one pose can be estimated in one inference.

In this work, we improve the PVNet architecture to choose the correct normalization parameters automatically. Class-adaptive instance (de)normalization (CLADE) \cite{Tan2020} selects object-specific parameters based on the semantic class of each pixel. The resulting network can easily be trained for multiple objects, the multi-object gap is narrowed, and all known objects in one image can be found during one inference without knowing their identity in advance. 

\paragraph{Local object tracking.}
\label{sec:pose_estimation}
Although pose estimation with CNNs has recently developed very rapidly, state-of-the-art results for model-based frame-to-frame tracking are still achieved with non-learning-based methods. For potentially non-textured objects and image-based tracking, existing approaches can roughly be separated into edge and region-based methods.

Region-based methods use either global \cite{Prisacariu2012} or temporary local color \cite{Tjaden2018,Zhong2019} histograms to separate an object from the background and optimize the pose to maximize the discrimination. They are best suitable for objects, which are distinct from the background, but easily fail for objects, which have a similar color to the background \cite{Sun2021}. 

Edges or contours are suitable visual cues for tracking non-textured objects. Based on the RAPID algorithm \cite{Harris2019}, 2D-3D correspondences are searched on scanlines perpendicular to the object contour. Extensions filter those correspondences with respect to the contour orientation \cite{Huang2020}, a global or local color histogram \cite{Seo2013,Wang2015,Huang2020} or consider multiple hypotheses per scanline. Other edge-based algorithms do not explicitly include point-to-point correspondences, but instead minimize a pixel-based distance metric directly on the intensity image \cite{Dong2020,Wang2019}. 

Similarly, analysis-by-synthesis-based methods \cite{Seibold2017,Gard2019} try to synthetically recreate the camera image with a rendered representation and minimize image distance by motion compensation with respect to the optical-flow constraint. A comparable representation between real and synthetic images has to be found, either by explicitly modelling scene or image-parameters such as motion blur \cite{Seibold2017}, or simplification, e.g.\ by using robust edge-images \cite{Gard2019}.

Our tracking algorithm combines correspondence and non-correspondence-based tracking, independent of color information, and is suitable for textured and non-textured objects. It bridges larger pose differences during RAPID-based iterations. A subsequent analysis-by-synthesis optimization makes fine adjustments and also accounts for inner edges.

\section{\uppercase{Registration Pipeline}}
\label{sec:method}

This section covers our multi-object tracking and identification pipeline. Our system should not need any photos of the real object before being able to recognize it. In an offline phase, renderings are generated as training data for our pose estimation CNN (\autoref{sec:data_generation}). The CNN constantly scans the video stream and estimates an initial pose for each visible, known object (\autoref{sec:neural_network}). After detection, a local refinement step maps the edges of the 3D model to the image edges (\autoref{sec:local_refinement}). A validation step (\autoref{sec:pose_validation}) evaluates the quality of these matches and either validates or invalidates the pose. For objects with validated poses, it is sufficient to proceed with local refinement for the next image.
  
\subsection{Data Generation}
\label{sec:data_generation}

The Input of the data generation step is a set of multiple 3D models to be detected and tracked. They may have similar shapes and may be untextured or from the same material but should not be rotationally symmetric. The models are centered at their 3D bounding box and $k$ keypoints, the object center and $k-1$ points on the object surface, are determined using the Farthest Point Sampling algorithm \cite{Peng2019}. 

We create a dataset consisting of synthetic renderings using NDDS \cite{To2018}, an Unreal Engine plugin for generating annotated training images. To address the domain gap, i.e.\ the different properties of real and synthetic images that affect the accuracy of a CNN trained on synthetic data only, we use domain randomization \cite{Tobin2017}. We render the objects in front of a random background, which is either a photo or a randomly generated procedural graphic. In half of the images, the objects are rendered over a flat surface to introduce shadows. Also, the position of light sources, the orientation and position of the object, the texture of the object, and the position of the camera are randomized. Randomly placed distractor objects introduce partial occlusion. In each image, only one target object is visible, which simplifies the recompilation of new datasets for different sets of objects.

\begin{figure}
\flushleft  
\begin{subfigure}{.31\columnwidth }
  \centering
  \includegraphics[width=\linewidth]{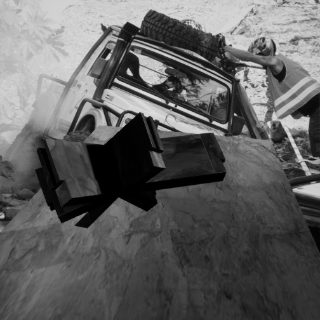}
\end{subfigure}
\hspace{.01\columnwidth}
\begin{subfigure}{.31\columnwidth }
  \centering
  \includegraphics[width=\linewidth]{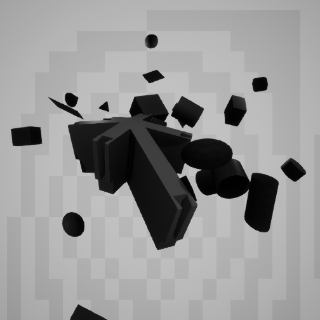}
\end{subfigure}
\hspace{.01\columnwidth}
\begin{subfigure}{.31\columnwidth }
  \centering
  \includegraphics[width=\linewidth]{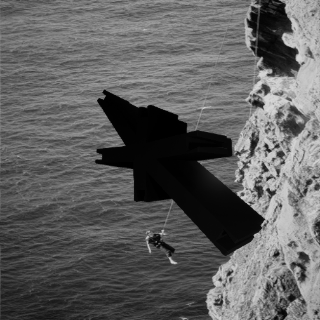}
\end{subfigure}

\begin{subfigure}{.31\columnwidth }
  \centering
  \vspace{0.1cm}
  \includegraphics[width=\linewidth]{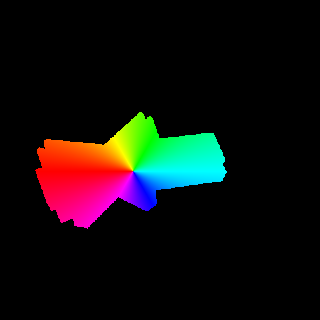}
\end{subfigure}
\hspace{.01\columnwidth}
\begin{subfigure}{.31\columnwidth }
  \centering
  \vspace{0.1cm}
  \includegraphics[width=\linewidth]{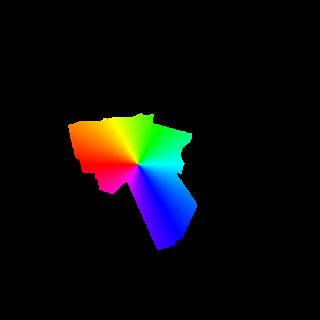}
\end{subfigure}
\hspace{.01\columnwidth}
\begin{subfigure}{.31\columnwidth }
  \centering
  \vspace{0.1cm}
  \includegraphics[width=\linewidth]{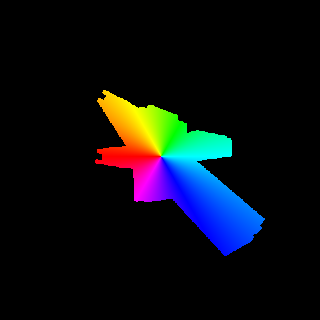}
\end{subfigure}
\caption{Synthetically rendered training images and the corresponding color-coded vector fields that, within the object mask, point to the 2D projection of a 3D keypoint.}
\label{fig:train}
\end{figure}

Contrary to other publications \cite{Hodan2019,Thalhammer2021}, we do not render near-photo-realistic images and instead deal with inaccurate pose estimates by refining the pose locally and filtering wrong pose estimates by our pose validation step. We further reduce the domain gap and focus on shape differences by using grayscale images only. For each training image, a mask of the visible part of the object, its pose, and the position of the keypoints in 3D and 2D space are stored (\autoref{fig:train}).

\subsection{6D Detection Network }
\label{sec:neural_network}
Our pose estimator first establishes 2D-3D correspondences with a correspondence estimation CNN. It predicts $n+1$ pixel-level masks for the background and $n$ known objects, as well as $k$ joint vector fields. The number of keypoints is the same for each object. In a vector field, two coordinate maps form 2D vectors pointing to the image location of the keypoint belonging to the object a pixel is assigned to in the semantic segmentation. 

As with PVNet \cite{Peng2019}, the intersections of randomly selected vector pairs within an object mask result in 2D location estimates that are validated with a RANSAC-based voting procedure. The object pose is found with a PnP algorithm. Contrary to them, the joint vector field reduces the number of output maps for the vector fields from  $2nk$ to $2k$ and is independent of the number of objects. This makes the network much easier and faster to train, reduces the required GPU memory during training, and the data transfer between GPU and CPU after inference. E.g.\ for 13 objects with 9 keypoints the number of output maps are reduced from 248 to 32. 

The following modifications are made to PVNet.
\begin{enumerate}
\item The semantic segmentation and the vector fields are predicted with two different decoders connected to the same encoder.
\item In the keypoint decoder, the batch normalization is replaced with a class-adaptive instance (de)normalization (CLADE) \cite{Tan2020}.
\item The estimated semantic segmentation is used as a side input for the CLADE layers to select object-specific weights with the Guided Sampling \cite{Tan2020} strategy, based on the class a pixel belongs to.
\end{enumerate}
The object-specific weights increase the capacity of the network for multi-object pose estimation. The spatial selection of those weights allows correspondences for multiple objects to be estimated with one inference. In the decoder blocks, the semantic mask is downsampled so that its size matches with the output of the. convolution.

\begin{figure}
\centering
  \includegraphics[width=\columnwidth]{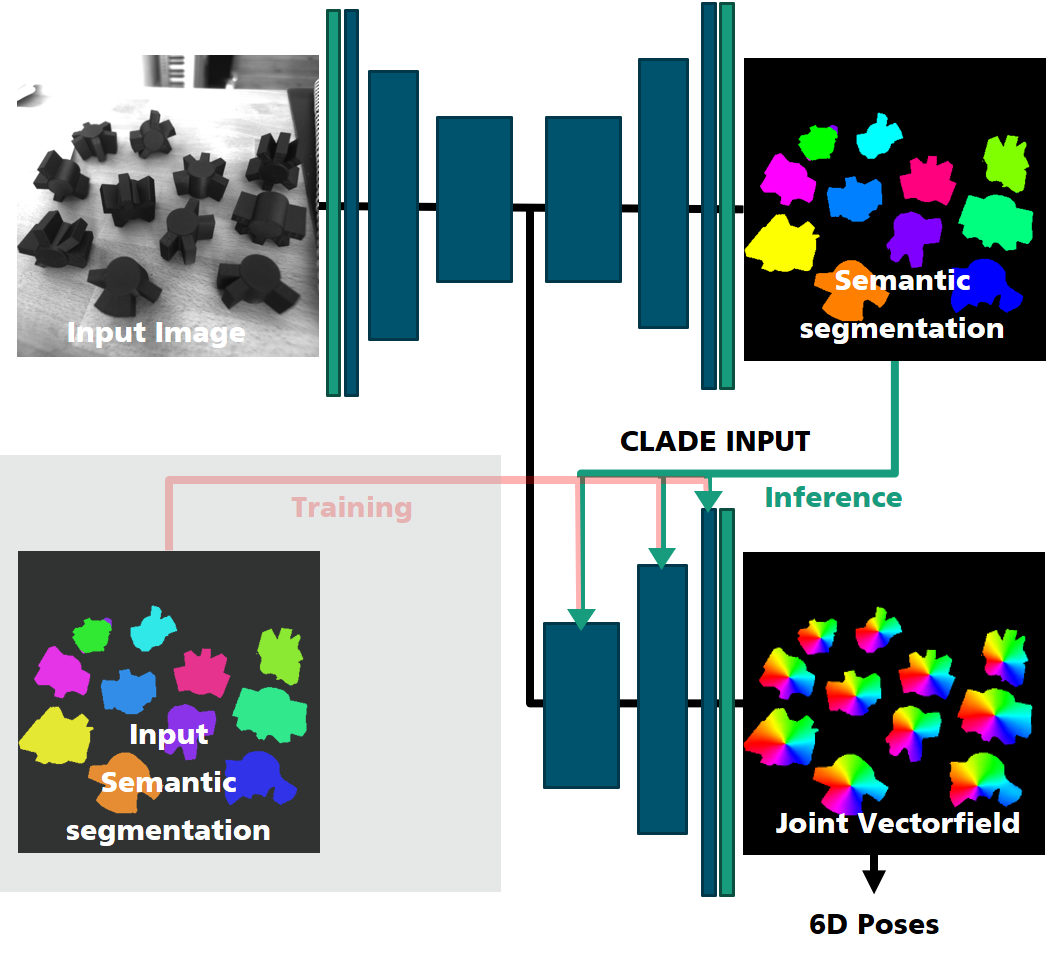}
 \caption{The input image is processed with a segmentation branch that guides the vector field prediction for 2D-3D correspondence estimation.}
 \label{fig:architecture}
\end{figure}

To achieve better convergence during training, the ground truth mask instead of the estimated mask is used as side input. During inference, the intermediate logits of the segmentation are normalized with a scaled softmax function with high temperature value. Our CNN architecture is visualized in \autoref{fig:architecture}.

\subsection{Local Refinement}
\label{sec:local_refinement}
Poses estimated by the CNN are not stable enough for AR use, as even small jitter significantly reduces visual quality. Furthermore, a domain gap between camera images and synthetic training images affects the accuracy of the tracking. Therefore, we apply an additional local analysis-by-synthesis refinement to stabilize the pose estimation. If possible, only the local refinement is used, since it can bridge the small frame-to-frame movements.

The refinement starts from an approximate pose given by $3\times3$ rotation matrix $\mathbf{R}$ and translation vector $\mathbf{t}$. This pose is either the output of the global CNN detection or the output of local refinement at time step $t-1$. A synthetic image $\hat{I}$ and a depth map $D$ are generated by an off-screen renderer using meshes of the detected objects. The goal is to find the pose offsets $\Delta \mathbf{t}$, $\Delta \mathbf{R}$ that compensate the difference between camera image $I$ and $\hat{I}$.

If $\mathbf{\hat{p}} = [p_x, p_y, p_z]^T$ is a 3D point on the model surface in $\hat{I}$, it is transformed to $\mathbf{p}$ and projected into the image point $\mathbf{x}$ with the intrinsic matrix $\mathbf{K}$ and a homogenization operation $\pi(\mathbf{p}) = [p_x/p_z, p_y/p_z]^T$.

\begin{equation}
\label{equ:projection}
\mathbf{p} = \Delta \mathbf{R} ( \mathbf{\hat{p}} - \mathbf{t} ) + \mathbf{t} + \Delta \mathbf{t}
\end{equation} 
\begin{equation}
\mathbf{x} = \pi(\mathbf{K} \mathbf{p})
\end{equation} 

As in \cite{Steinbach2001}, under small motion assumption, a linearized rotation matrix with three unknown parameters $\Delta \mathbf{r} = [\Delta r_x, \Delta r_y, \Delta r_z]^T$ is used and the displacement error is expressed as a linear equation using first order Taylor expansion. Our optimization algorithm obtains a two-stage structure of two consecutively solved minimization problems. The usage of an image pyramid and an iteratively reweighted least squares (IRLS) scheme \cite{Zhang1997} stabilize convergence and reduce the influence of outliers. 

\subsubsection{Contour-based Optimization}
Inspired by other work \cite{Huang2020,Harris2019}, we implement an edge-based registration algorithm. The depth map $D$ allows generating a silhouette mask from which $m$ edge points $\mathbf{e}_i$ and the corresponding 3D points $\mathbf{\hat{p}}^e_i$ are extracted. 

We extract match hypotheses along $m$ scanlines $l_i$, along the unit vector $\mathbf{s}_i$ perpendicular to the projected contour at a given point. We sample $I$ along $l_i$ and convolve each sample with pre-computed $5\times5$ rotated Sobel kernels to extract edges with similar orientation to the projected edge. All locations where the convolved value is a local maximum along the scanline and larger than a threshold $t_e$, are stored as edge hypothesis points $\mathbf{h}_{i,j}$. We minimize the error function
\begin{equation}
E(\Delta \mathbf{R}, \Delta \mathbf{t}) = \sum_{i=0}^m \omega(r_i)({\mathbf{s}_i}^T (\mathbf{h}_i - \pi(\mathbf{K} \mathbf{p}^e_i))) \,
\end{equation} 
whereby the pose delta transforms $\mathbf{\hat{p}}^e_i$ to $\mathbf{p}^e_i$ (\autoref{equ:projection}) and $\mathbf{h}_{i}$ is the hypothesis with the smallest spatial distance to the observed contour point. By using the Taylor approximation of the displacement error, we solve an overdetermined linear equation system for the pose parameters. The weighting function $\omega(x) = 1 / \sqrt{x^2+ \epsilon^2}$ applies the robust Charbonnier penalty \cite{Sun2010} with $\epsilon = 0.001$ on the hypothesis residual $r_i$ from the previous IRLS iteration. Two or three weight updates stabilize the estimation against outliers. 

Each pyramid stage consists of four to six repetitions. The off-screen renderer and the hypothesis selection are only executed initially. Consecutively, only the extracted 3D contour is transformed. Moreover, on the smallest pyramid level, we only solve for translation and rotation within the 2D image plane.

\subsubsection{Dense Refinement}
The contour-based optimization may be affected by mismatches along the scanlines. We use a subsequent dense optimization minimizing the image distance with respect to the general optical flow equation \cite{Horn1981}
\begin{equation}
\label{equ:optical flow}
\frac{\partial \hat I}{\partial \hat X_1}\mathbf{u_m} + \frac{\partial \hat I}{\partial \hat Ys_1}\mathbf{v_m} \approx \hat I - I
\end{equation}
which is reformulated to be a linear equation system. This system is solvable with respect to the unknown pose parameters using the coefficients $\mathbf{a}$ from \cite{Steinbach2001}.
\begin{equation}
\mathbf{a} \begin{pmatrix} \Delta\mathbf{r} \\ \Delta\mathbf{t}  \end{pmatrix} = \hat I - I
\end{equation}

Every pixel inside the silhouette of the rendered object results in one equation for the iterative reweighted least squares solution.

Adaptive thresholding \cite{Gard2018,Gard2019} makes the rendered image and the camera image comparable. First, a thresholded Sobel filter extracts edges in the rendered image, then the threshold for the camera image is adapted over a 2D grid to reproduce a similar distribution of edge and non-edge pixels in each grid cell. This also reduces the influence of illumination.

Both edge images are smoothed with a box filter to introduce smooth gradients around the detected edges before minimizing the image distance. The optimization is suitable for textured and non-textured objects. Due to the application of simple shading in the off-screen renderer, sharp geometric edges will be visible in the Sobel image.

Since all operations are applied pixel-wise, the processing can be executed on the GPU by custom compute-shaders. Also, the equation coefficients are combined on the GPU. The least squares problem is formulated in $\mathbf{Ax} = \mathbf{b}$ form and each iteration only needs a transfer of the symmetric $6\times6$ matrix  $\mathbf{A}^T\mathbf{A}$ and the $6\times1$ vector $\mathbf{A}^T\mathbf{b}$ between GPU and CPU, where the system is solved.

\subsection{Pose Validation}
\label{sec:pose_validation}
In a single-camera system, objects with small geometric differences can lead to ambiguities that have to be filtered reliably. A pose validation filters wrong pose estimates due to drift or false-positive detections.

\paragraph{Detection validation.}
The RANSAC-based voting \cite{Peng2019} during detection needs a minimum number of positive samples (e.g.\ $25$) to validate a 2D keypoint hypothesis. After the initial pose estimation, the reprojection error must be smaller than a threshold (12 pixels in our experiments) for at least 4 points.

\paragraph{Refinement validation.}
The initial pose can still be coarse, so pose validation including acceptance or rejection is performed on the refined poses. It requires the projected contour of the object to match with the image content. We find the error of the RAPID-inspired local registration suitable to measure this. 
Here, $e_{IRLS}$ is the mean residual value in the last iteration in the lowest pyramid level, $e_{Dist}$ is the mean distance between points on the projected contour and the closest correspondence hypothesis, and $e_{valid}$ is the ratio of the number of scanlines to the number of scanlines on which a correspondence hypothesis was found. We define the edge matching score as
\begin{equation}
e_{edge} = e_{IRLS} \times e_{Dist} \times e_{valid} \quad.
\end{equation} 

An initial pose is refined and $e_{edge}$ of the first frame is stored as initial error $e_{init}$. If it is smaller than a threshold $\tilde e_{init}$ an initial pose is accepted. Otherwise, we keep the object as a candidate and further refine the pose locally for the next frames as long as $e_{edge}$ decreases. If the error increases compared to the last frame, a detection is discarded.

Then, $e_{edge}$ is monitored continuously and the running mean average $\hat e_{edge}$ is updated as long as the object remains valid. As long as $e_{edge}$ is smaller than a maximum value $\tilde e_{max}$ and $\hat e_{edge} \times f$, where $f$ is a predefined factor, an object keeps its validity. 
If not, the object is in a borderline state. The pose is not updated and a mismatch counter increases 
for every following invalid frame until either the pose becomes valid again or two invalid frames appear in a row, which sends the object to an uninitialized state.

\section{\uppercase{Implementation Details}}

\subsection{CNN Training and Architecture}
Our training dataset consists of $\sim$15000 synthetic images per object. Our CNN is trained for 125 epochs using the ADAM optimizer. The initial learning rate of $0.002$ is divided by two every 25 epochs. 
The backbone of our network provides features for both decoders and is connected via skip connections to both of them. It is a pretrained ResNet-18 \cite{He2016} obtaining the same modifications as in \cite{Peng2019}.
In the keypoint decoder, each convolution is followed by a CLADE layer. The CLADE layers extend the CNN by 1024 trainable parameters per object. In total, the number of trainable weights increases by about 14\% by the second decoder.

Training images have a size of $320^2$. The images are augmented by random contrast and brightness, additional normally distributed noise, and random rotation and translation variations. The input images are grayscale. To use the pretrained weights of ImageNet, the intensities are stacked to form a three-channel image. We use differentiable proxy voting loss (DVPL) \cite{Yu2020} and smooth l1-loss to learn the vector field. We use softmax cross-entropy loss to learn semantic segmentation.

During training, we input the ground truth segmentation to the CLADE layers of the vector field decoder. The numbers given for PVNet are computed with our Tensorflow port of the original code.

\subsection{CNN Inference and Live Tracking}
\label{sec:live_tracking}

The inference in our tracking software uses the OpenCV DNN module with its CUDA extension. Initial poses are estimated from 2D-3D correspondences with EPNP \cite{Lepetit2009}.

The refinement exploits an off-screen OpenGL renderer to compute the intermediate images. While the contour-based part is fast on CPU, the dense refinement relies on custom GLSL compute shaders for all image processing and the formation of the equation system. Only solving the equation system is done on the CPU, minimizing data transfer. 

The resolution of all test images is $1024^2$, input size of the CNN is $400^2$. The local tracking uses three pyramid levels. The length of the scanlines is 15.  

Two tracking modes have been implemented. In \textit{Close-Range} mode, the subsampled camera image is passed to the detection network. Multiple objects are searched at the same time and objects are tracked independently. The distance between the objects and the camera matches the training data. 

In \textit{Far-Range} mode, the distance between object and camera is much larger than in the training data. We propose to pass image patches to the CNN in which the size of the object relative to the patch size matches the median size of the objects in the training images. 

\section{\uppercase{Experiments}}
\label{sec:eval}
\begin{figure}[h]  
  \centering
  \includegraphics[width=\columnwidth]{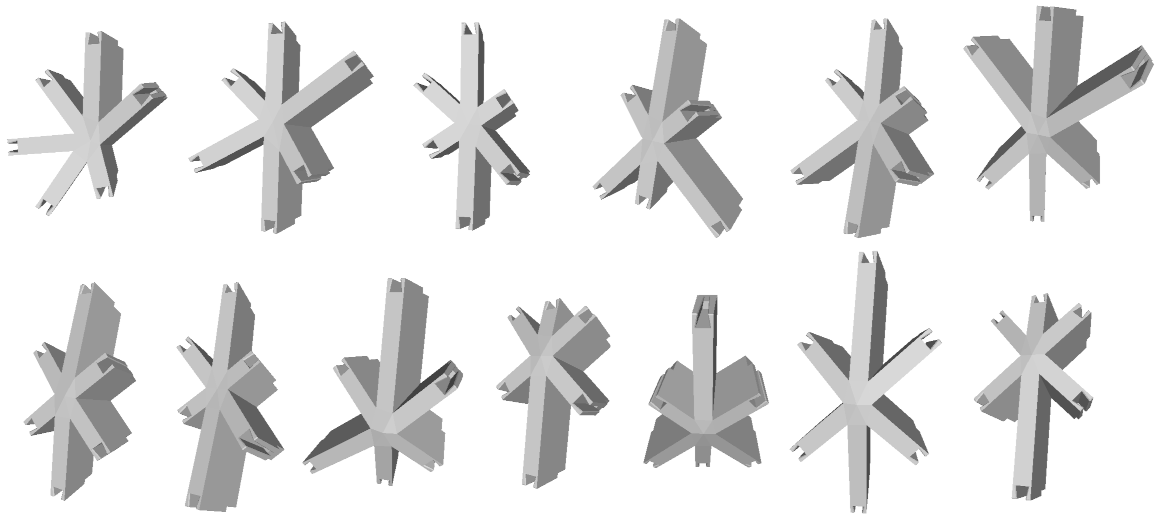}
\caption{Renderings of the 13 objects (\textit{I01} -\textit{I13}) used in our experiments, numbered from top left to bottom right.}
\label{fig:objects}
\end{figure}

We evaluate our registration pipeline with synthetic data first and then describe general observations on real data in the scenario of AR-guided construction. 
Within that example use case, the evaluation models are part of a miniaturized model of a grid shell facade, consisting of 13 node elements (\autoref{fig:objects}) and 42 connector sticks. The node elements are particularly interesting since they all have similar but not identical shapes and geometric ambiguities can arise easily.

The \textbf{2D projection metric} \cite{Brachmann2016} is used to judge the detection accuracy. The vertices of the models are projected into the image with the estimated pose and the ground-truth pose. A pose is correct if the average distance is smaller than 5 pixels and the percentage of correct estimates is listed. 

\subsection{Benefits of the Multi-Object Detection Network}

\label{sec:synth_1}

\begin{figure}[t]
\flushleft  
\begin{subfigure}{.48\columnwidth }
  \centering
  \includegraphics[width=\linewidth]{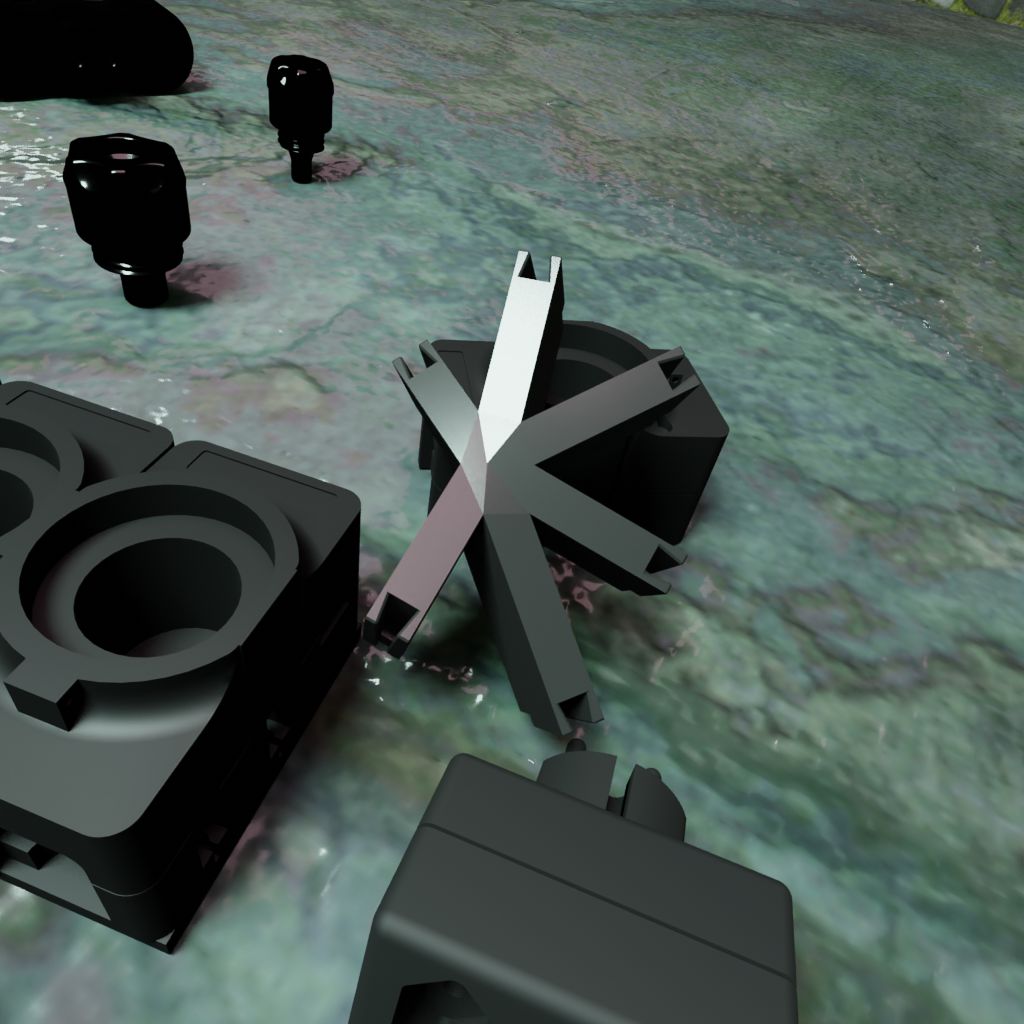}
\end{subfigure}
\hspace{.01\columnwidth}
\begin{subfigure}{.48\columnwidth }
  \centering
  \includegraphics[width=\linewidth]{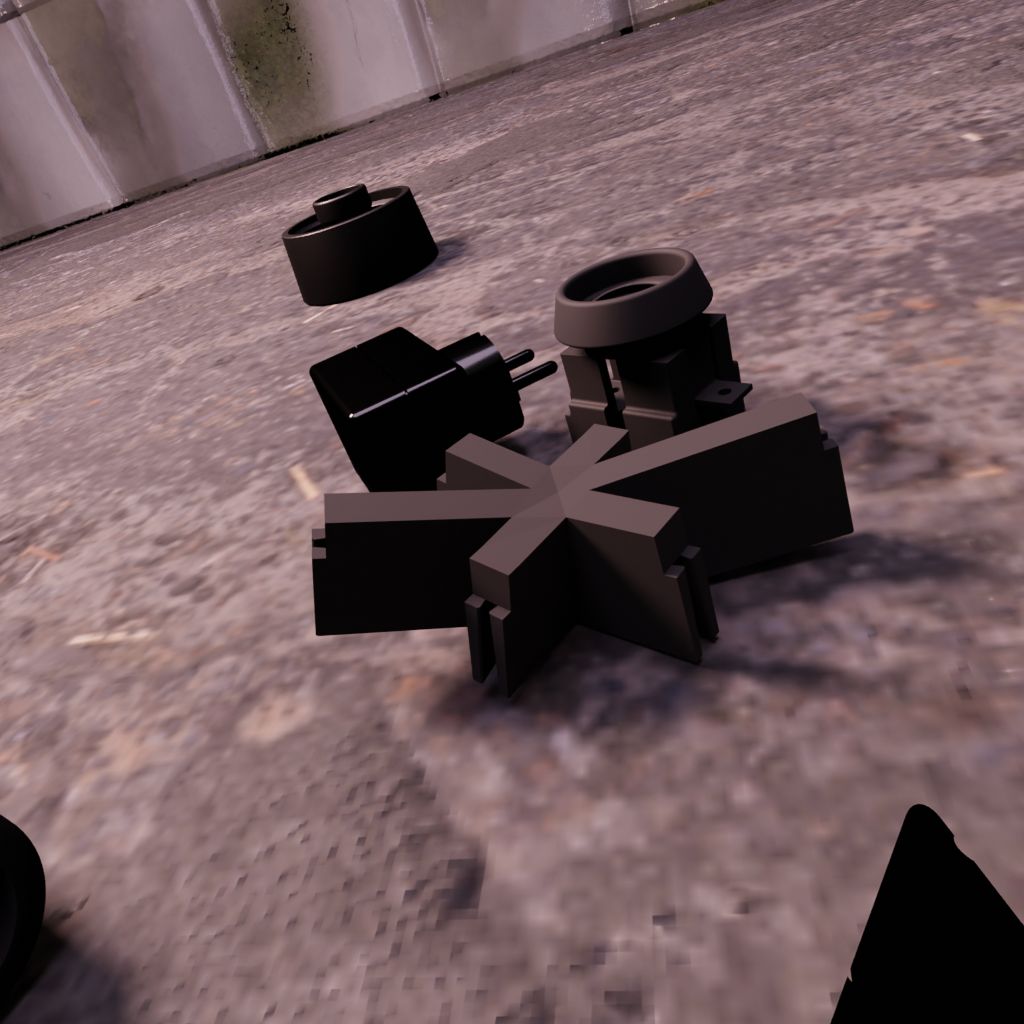}
\end{subfigure}
\caption{Two images from our synthetic evaluation set.}
\label{fig:eval}
\end{figure}

The objects in commonly used pose estimation datasets, like Linemod \cite{Hinterstoisser2012} or YCB Video \cite{Xiang2018}, are clearly identifiable by varying color and shape. We focus on the benefit of our multi-object model to differentiate between similar objects. 

Our evaluation dataset contains 200 images per object. Images are generated synthetically but a different render engine is used to introduce a domain gap between training and testing data.
The synthetic data generator BlenderProc \cite{Denninger2019} based on Blender allows rendering nearly-photo realistic images. Two examples are shown in \autoref{fig:eval}. In each image, the object is placed on a flat surface with varying texture and is surrounded by distractor objects. The camera, object, and light source poses vary between the images. 
First, we compare three multi-object models:

\begin{enumerate}
\item \textbf{PV-M} uses the trivial multi-object extension of PVNet. With every added object, 19 channels are added to the network output. 
\item \textbf{PV-M-C} is a modification of the baseline architecture that outputs joint vector fields.
\item \textbf{CLADE-PV} uses our proposed network structure, as described in \autoref{sec:neural_network}.
\end{enumerate}

\autoref{table:synth_1} lists the results for the different network structures with respect to the 13 objects. Our model improves the detection accuracy compared to the baseline model by a large margin. It is important to note that \textbf{CLADE-PV} requires much less GPU memory than \textbf{PV-M} during training. With \textbf{CLADE-PV}, seven times larger batch sizes were possible on the same hardware, resulting in much faster training. The \textbf{PV-M-C} variant was designed to be trainable with the same batch size as \textbf{CLADE-PV} but performed worst.

In the next experiment (\autoref{tab:single}), we compare multi and single-object models. We train two separate PVNet configurations for six different objects, differing in the training data used. \textbf{PV-S} sees only the images of one object. It produces false-positive detections most of the time for all other objects. \textbf{PV-S+} also sees the same amount of randomly selected images of other objects to learn to distinguish its own object from the others. Still, if the identity of the objects present in the image is unknown, all networks have to be tested with the input image which will result in computational overhead. In conclusion, on average the single-object models perform slightly better than the multi-object competitors, confirming the multi-object gap, but also worse than our proposed modified network. 

\begin{table}
\caption{\label{table:synth_1}Accuracies of our method and the baseline methods on our evaluation dataset using \textbf{2D projection} metric.}
\begin{center}
 \begin{tabular}{|c |c |c | c |} 
 \hline
 Method & PV-M & PV-M-C & CLADE-PV  \\ 
  &  &  & (ours) \\ 
  \hline
 $I01$ &  75.0 & 60.0 & \textbf{79.0}\\ 
 $I02$ &  78.0 & 54.5 & \textbf{79.0} \\ 
 $I03$ &  71.0 & 54.0 & \textbf{76.5} \\ 
 $I04$ &  82.5 & 68.0 & \textbf{86.5} \\ 
 $I05$ &  93.5 & 62.5 & \textbf{95.5} \\ 
 $I06$ &  66.5 & 41.5 & \textbf{85.0} \\ 
 $I07$ &  80.0 & 70.0 & \textbf{90.0} \\ 
 $I08$ &  73.3 & 54.2 & \textbf{82.4} \\ 
 $I09$ &  53.5 & 49.5 & \textbf{87.0} \\ 
 $I10$ &  75.0 & 80.0 & \textbf{94.0} \\ 
 $I11$ &  85.0 & 68.5 & \textbf{91.5} \\ 
 $I12$ &  69.0 & 53.5 & \textbf{77.0} \\ 
 $I13$ &  84.0 & 72.5 & \textbf{90.0} \\ 
 \hline
  \textbf{Avg}. &  75.9 & 60.6  & \textbf{85.6}\\  
 \hline
\end{tabular}
\end{center}
\end{table}
                                                     
\begin{table}
\caption{\label{tab:single} Comparison of single-object models to our multi-object model using \textbf{2D projection} metric. }
\begin{center}
 \begin{tabular}{|c |c |c | c |} 
 \hline
 Method & PV-S & PV-S+ & CLADE-PV  \\ 
  &  &  & (ours) \\ 
  \hline
 $I01$ &   76.5 & \textbf{80.5} & 79.0\\ 
 $I02$ &  71.0 & 70.0 & \textbf{79.0} \\ 
 $I03$ &  63.5 & 61.0 & \textbf{76.5} \\ 
 $I04$ &  84.5 & 77.0 & \textbf{86.5} \\ 
 $I05$ &  93.5 & 87.0 & \textbf{95.5} \\ 
 $I06$ &  \textbf{86.5} & 78.0 & 85.0 \\ 
 \hline
  \textbf{Avg}. &  79.3 & 75.6  & \textbf{83.6}\\  
 \hline
\end{tabular}
\end{center}
\end{table} 

\subsection{Benefits of the Combined Approach}
\paragraph{Local refinement.}
We first show, how the pose refinement stabilizes the estimations of the CNN. The raw initial poses are refined using the algorithm from \autoref{sec:local_refinement}. While the CNN is limited by the input resolution, the refinement uses the full image resolution. 
                 
In \autoref{tab:refinement}, we list the percentage of valid frames regarding the 2D projection metric with respect to the high resolution and use thresholds of 5 (\textbf{2D$_{<5}$}) and 1 (\textbf{2D$_{<1}$}) pixel. Also, the average rotational \cite{Tjaden2018} error ($\varnothing \mathbf{R}$) over all valid frames (5-pixel threshold), and the percentage of estimates where the projection error is reduced ($\mathbf{P}$\textbf{++}) are listed. The results for \textit{init.} correspond to the result from the last section but refer to the higher resolution.
We compare using contour-based (\textbf{S1}) or dense refinement (\textbf{S2}) alone with the combined approach (\textbf{S1+S2}), in which the two processes are alternated in the image pyramid. 

\begin{table}
\begin{center}
\caption{\label{tab:refinement} Improvement of refinement on initial poses. }
 \begin{tabular}{|l | c ||c |c || c | c |} 
 \hline
  Method& Iter. & 2D$_{<5}$ & 2D$_{<1}$ &$\varnothing R$  & $P$++\\ 
  \hline
  init. & $-$ & 26.4 & 0.0 & 3.50\degree & $-$\\
   \hline 
 S1 & 1 & 76.5 & 45.7 & 1.44\degree & 92.3  \\
 S2 & 1 & 73.8 & 27.3 & 1.54\degree & \textbf{97.8}  \\ 
 S1+2 & 1& 84.3 & 64.7 & 0.74\degree & 97.1  \\
 \hline  
 S1 & 3 & 78.7 & 52.9 & 1.20\degree & 90.0  \\ 
 S2 & 3 & 83.7 & 57.8 & 0.85\degree & 97.7  \\ 
 S1+2 & 3& \textbf{86.5} & \textbf{73.9} & \textbf{0.59\degree} & 96.5  \\  
 \hline
\end{tabular}
\end{center}
\end{table} 
We see that one iteration through the image pyramid (\textbf{Iter.}) of \textbf{S1+S2} improves the accuracy more than three iterations of \textbf{S1} or \textbf{S2} alone. \textbf{S1} converges fast, in the first iteration and can potentially bridge larger gaps, while \textbf{S2} is more likely to converge in the right direction (see $\mathbf{P}$\textbf{++}). \textbf{S1+S2} joins both advantages, which is beneficial for real-time systems. More iterations further improve the accuracy.

\paragraph{Pose validation.}
During the pose refinement, the error value $e_{edge}$ is the criteria for acceptance or rejection of poses. Ideally, it is larger than $\tilde e_{init}$ if the estimated pose is wrong and smaller than $\tilde e_{init}$ if the estimated pose is correct. 

\autoref{tab:validation} shows the influence of $\tilde e_{init}$.
The whole dataset is passed through the registration pipeline (\autoref{sec:method}) and every object is searched in every image. Three iterations of refinement are allowed.
We list the following percentages: 
\begin {enumerate*}[label=\arabic*)]
\item  correct estimates (5-pixel threshold) correctly validated (\textbf{2D$_{Proj}$}) (relative to amount of images),
\item correct estimates correctly validated (\textbf{C$_{OK}$}) (relative to amount of correct detections), 
\item wrong estimates correctly declined (\textbf{F$_{OK}$}) (relative to amount of incorrect detections),
\item wrong estimates correctly declined using ADD \cite{Hinterstoisser2012} to judge pose quality (\textbf{F$_{OK_{ADD}}$}) (relative to amount of incorrect detections),
\item  false-positive detection of absent object incorrectly validated (\textbf{FP}) (relative to amount of images).
\end {enumerate*}

\begin{table} [h!]
\caption{\label{tab:validation} Pose validation results with different initialization thresholds. }
\begin{center}
 \begin{tabular}{|l |c |c |c | c | c |} 
 \hline
  $\tilde e_{init}$ & 2D$_{Proj}$  & C$_{OK}$ & F$_{OK}$ & F$_{OK_{ADD}}$& FP \\ 
  \hline
 0.08  & 80.4 & 94.1 & 95.2 & 99.5 & 0.12 \\  
 0.10  & 83.2 & 96.7 & 92.6 & 98.6  & 0.27\\ 
   \hline
 \textbf{0.12}  & 85.4 & 98.7 & 89.3 & 98.6 & 0.46 \\ 
   \hline
 0.14  & 85.5 & 99.1 & 85.5 & 97.7  & 0.81\\  
 0.16  & 85.6 & 99.6 & 77.5 & 94.7  & 0.86\\  
 \hline
\end{tabular}
\end{center}
\end{table} 
The choice of $\tilde e_{init}$ is a trade-off, between possibly declining correct estimates, if too low, or possibly accepting false detections, if too high. For our test set, $\tilde e_{init}=0.12$ is a good choice. \textbf{2D$_{Proj}$} decreases quickly for smaller values and \textbf{F$_{OK}$} decreases quickly for larger values. Furthermore, \textbf{F$_{OK_{ADD}}$} confirms that most of the drastically wrong poses are filtered correctly.

False-positively detected objects often provide strong perspective shape ambiguities, since not only the wrong object is detected, but also its projected edges overlap with the image. An interesting observation is that the CNN partitions the segmentation output between multiple candidates, when unclear about the object identity (\autoref{fig:ambiguity}). This may result in multiple identity/pose estimates for a single object. 
If a correct pose cannot be verified, it is likely that a large part of the contour is covered or not visible. False pose detections of present objects can result from false initial estimates converging to local minima, with large edge overlap. In both cases, a small object or camera movement is often sufficient in AR applications with continuous video to find a starting position that converges correctly.

\subsection{Benefits in Real Video Sequences}
\label{sec:qualitative}
While previous experiments justified the different components of our system under domain gap, we show the applicability to real-world data on captured sequences. The recordings come from a live AR demonstrator guiding a construction scenario in two phases. First, the objects are sorted in the right order (\textit{Close-Range} mode), then they are mounted in that order (\textit{Far-Range} mode). Videos showing the sequences including overlays are part of the supplementary material.

\begin{table*}
\caption{\label{tab:squence} Evaluation of real-data sequences. For each object, the number of frames with edge error smaller $\tilde e_{init}=0.12$ is listed. Only frames where interaction with the object (movement or occlusion) happens are considered.}
\begin{center}
 \begin{tabular}{|c | c || c |c | c | c | c | c | c| c |} 
 \hline
  Seq.& Method & valid & $I01$ & $I04$ &$I06$ & $I08$& $I10$ & $I11$ & FP \\ 
  \hline
 $1$& Init.+Ref. & $52.7\%$ & 6/34 & 24/154 & 17/64 & 447/646& 156/376 & 72/96 & 1/1215 \\
 $1$& Init.+Ref.+Valid. & $87.5\%$ & 16/34 & 138/154 & 29/64 & 614/646& 313/376 & 89/96 & 3/1215 \\
  \hline 
 $2$& Init.+Ref. & $78.0\%$ & 371/501 & $-$ & 311/373 & $-$ & $-$ & $-$ & 0/727 \\
 $2$& Init.+Ref.+Valid. & $90.4\%$ & 460/501 & $-$ & 330/373 & $-$ & $-$ & $-$ & 0/727 \\ 
\hline 
\end{tabular}
\end{center}
\end{table*}

\begin{figure*}
\centering  
\begin{subfigure}{.32\linewidth }
  \centering
  \includegraphics[width=\linewidth]{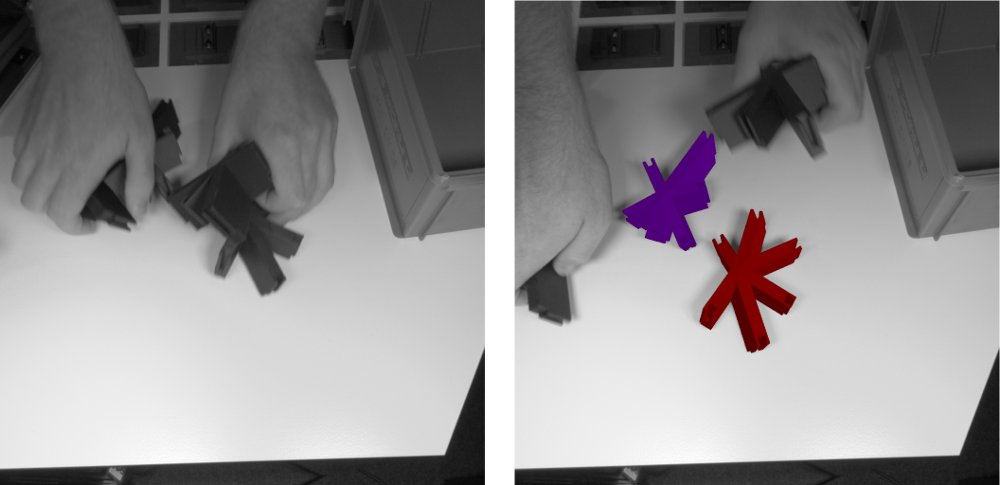}
  \caption{appearing objects}
  \label{fig:appear}
\end{subfigure}
\hspace{.005\linewidth}
\begin{subfigure}{.32\linewidth }
  \centering
  \includegraphics[width=\linewidth]{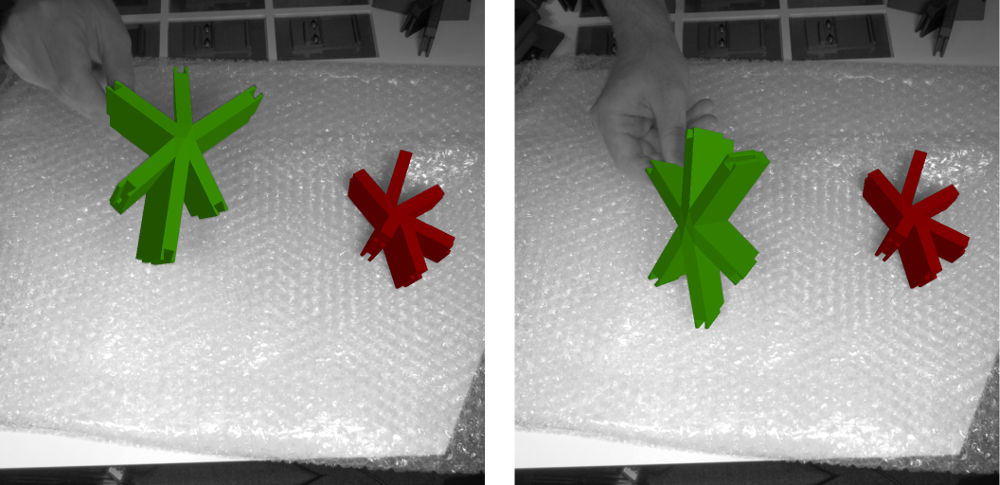}
  \caption{movement over textured background}
    \label{fig:bubblewrap}
\end{subfigure}
\hspace{.005\linewidth}
\begin{subfigure}{.32\linewidth }
  \centering
  \includegraphics[width=\linewidth]{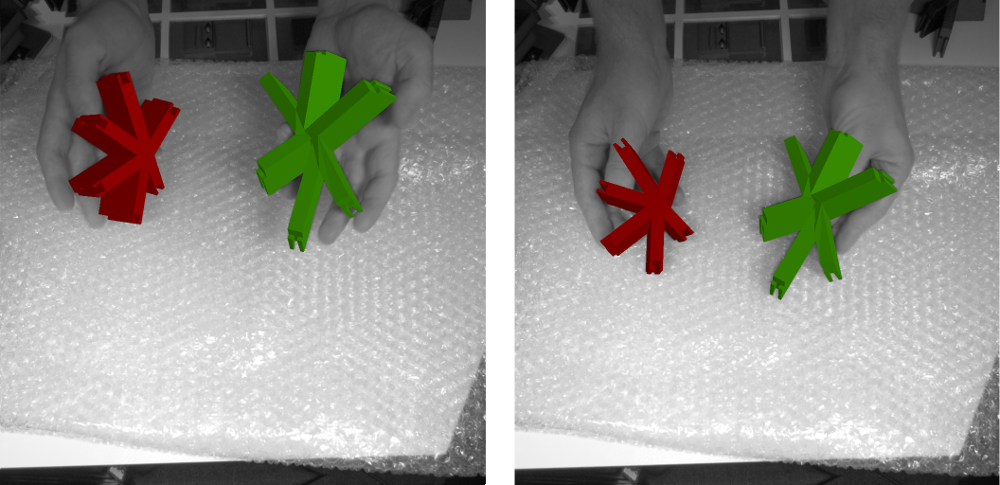}
  \caption{movement on hand palm}
  \label{fig:hand_palm}
\end{subfigure}
\centering  
\captionsetup{justification=centering}
\caption{Tracking and detection of objects in different situations: (a) is from \textbf{Seq. 1}, (b) and (c) are from \textbf{Seq. 2}.}
\label{fig:situations}
\end{figure*}

\paragraph{Close-Range mode.}
We exemplary demonstrate our pipeline in two sequences, captured with 25 fps and $1024^2$ pixels with an industrial camera. While \textbf{Seq. 1} has a white table background, \textbf{Seq. 2} has a textured bubble wrap background with stronger gradients and light reflections. 
Multiple randomly picked objects are moved in front of the camera by hand. 
The sequences depict fast movements, indirect movements, simultaneous movement of multiple objects, occlusion, and disappearing and reappearing objects.

\begin{figure}
\centering  
  \includegraphics[width=.40\linewidth]{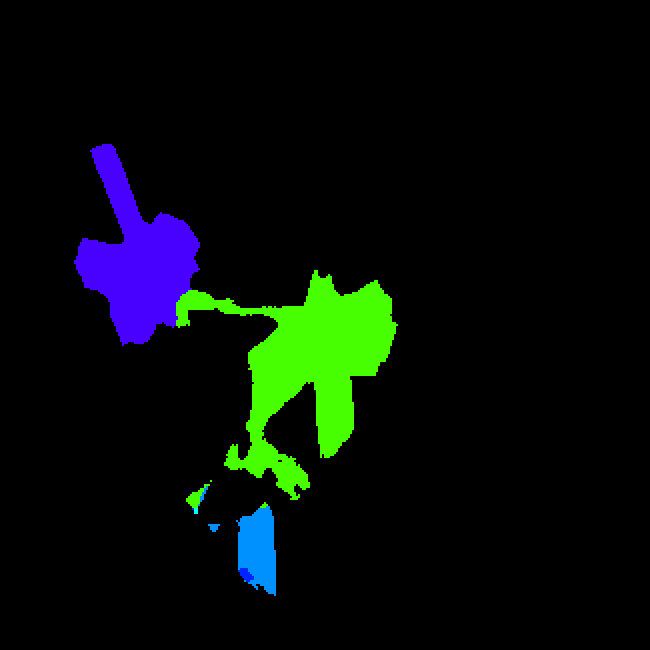}
  \hspace{.05\linewidth}
  \includegraphics[width=.40\linewidth]{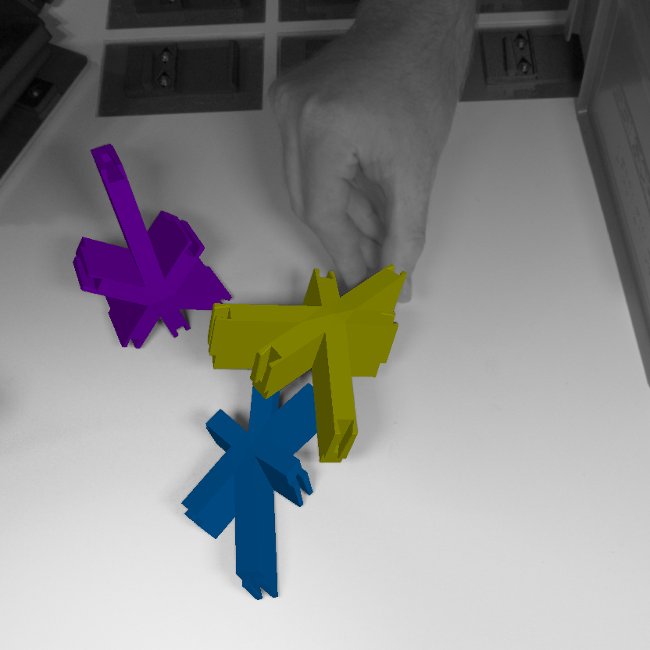}\\
\vspace{0.2cm}
  \includegraphics[width=.40\linewidth]{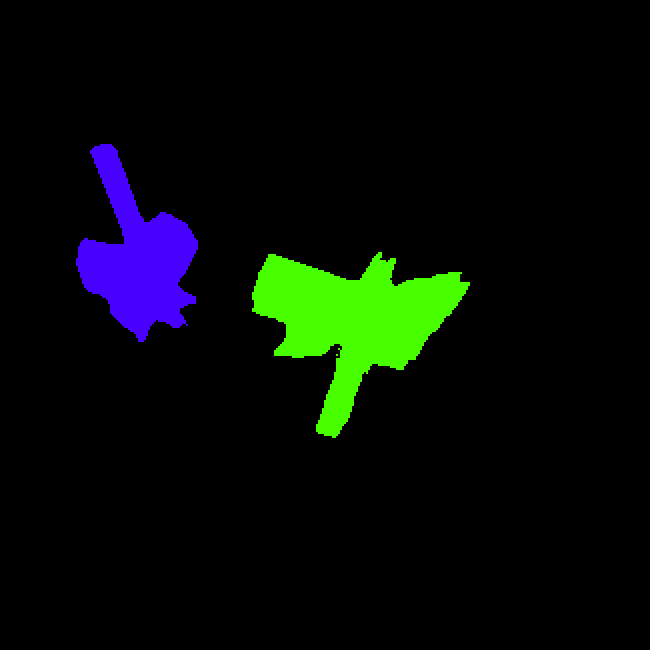}
    \hspace{.05\linewidth}
  \includegraphics[width=.4\linewidth]{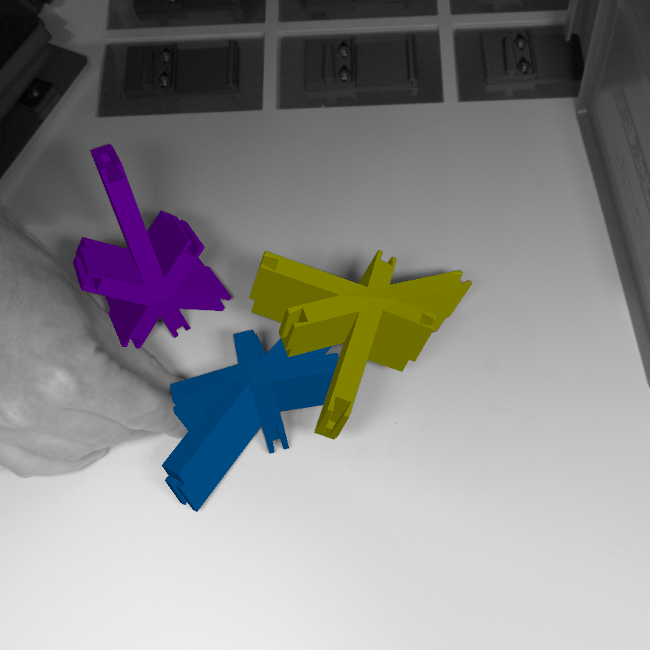}
  \caption{Local tracking remains valid (right), while the CNN (semantic segmentation, left) would be unreliable.}
\label{fig:indirect}
\end{figure}

To evaluate the tracking quality, we apply the multi-object tracking on the sequences and decide whether to refine or reinitialize individual objects for the next frame based on the validation criteria (\autoref{sec:pose_validation}). The result confirms our choice of $\tilde e_{init}=0.12$. Objects with $e_{edge}<\tilde e_{init}$ provide a nearly pixel-accurate overlay. The local and global registration benefit from each other in multiple ways:
\begin {enumerate}[label=\arabic*)]
\item  During training, the network has never seen multiple objects in an image; during testing, the network is able to estimate the poses of multiple objects simultaneously. Accuracy degradations are compensated by local refinement (\autoref{fig:appear}).
\item Occlusions between detectable objects were not seen during training, but even if the network is unable to find the occluded object, tracking will continue as long as local tracking remains valid (\autoref{fig:indirect}). 
\item Fast movements or heavy occlusions possibly stop the local tracking, but it continues as soon as the object is clearly visible again (\autoref{fig:tracking2}).
\item False-positive detections of the network are suppressed since the projected contour does not match image content well enough. 
\end {enumerate} 

The textured background in \textbf{Seq. 2} has no impact on the tracking accuracy (\autoref{fig:bubblewrap}). In the course of the sequence, the objects are also moved on the palm of a hand, whereby tracking as well as initialization also succeed in that situation (\autoref{fig:hand_palm}).

\autoref{tab:squence} shows quantitative results. For each object, we first count the number of frames in which it is either in motion or partially but no more than half (three out of six connectors are visible), occluded by a moving hand/object. We compare two configurations, one using the CNN and refinement independent of the previous pose for every frame (\textbf{Init.+Ref.}), and one dynamically selecting if the CNN is needed by pose validation (\textbf{Init.+Ref.+Valid.}). It can be seen that the number of correctly validated frames increases by the latter since the knowledge of the previous pose is often more meaningful than the result of the CNN. Nevertheless, on average more than half of the frames could be used for reinitializing local tracking. Fairly low values for $I01$ and $I06$ result from fast movements with motion blur, possibly under occlusion.

\begin{figure}
\centering  
  \includegraphics[width=.22\linewidth]{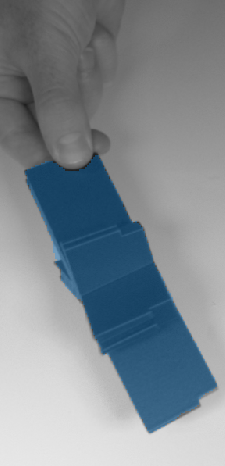}
  \hspace{.03\linewidth}
  \includegraphics[width=.22\linewidth]{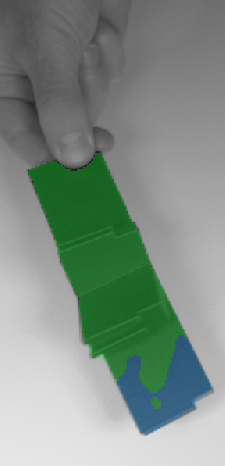}
  \includegraphics[width=.22\linewidth]{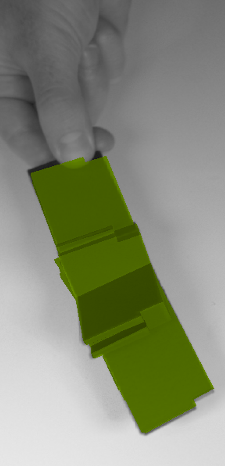}
  \includegraphics[width=.22\linewidth]{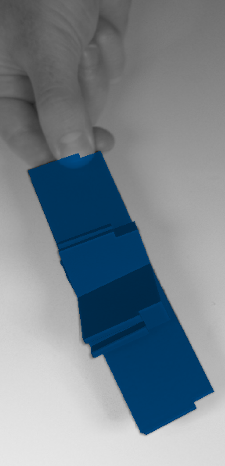}
  \caption{Geometric ambiguity: For a frame (left) the object can be identified (semitransparent mask overlay), while for the next frame (2nd) the mask is partitioned in two parts, resulting into two possible object poses (f.l.t.r.).}
\label{fig:ambiguity}
\end{figure}

In \textbf{Seq. 1}, ambiguous appearance of an object during rotation triggers a false-positive detection (\autoref{fig:ambiguity}). While the network is sure of the correct classification in the next frame, local tracking continues at first and only fails after a few frames if the projected contour deviates too much from the image content.

\begin{figure*}[t]
\centering  
\captionsetup[subfigure]{labelformat=empty}
\begin{subfigure}{.23\linewidth }
  \centering
  \includegraphics[width=\linewidth]{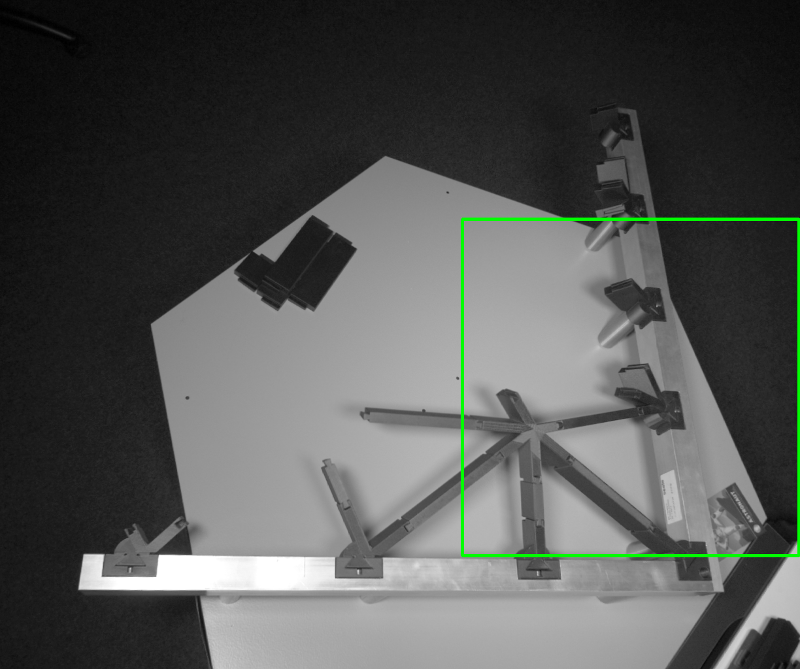}
\end{subfigure}
\hspace{.005\linewidth}
\begin{subfigure}{.23\linewidth }
  \centering
  \includegraphics[width=\linewidth]{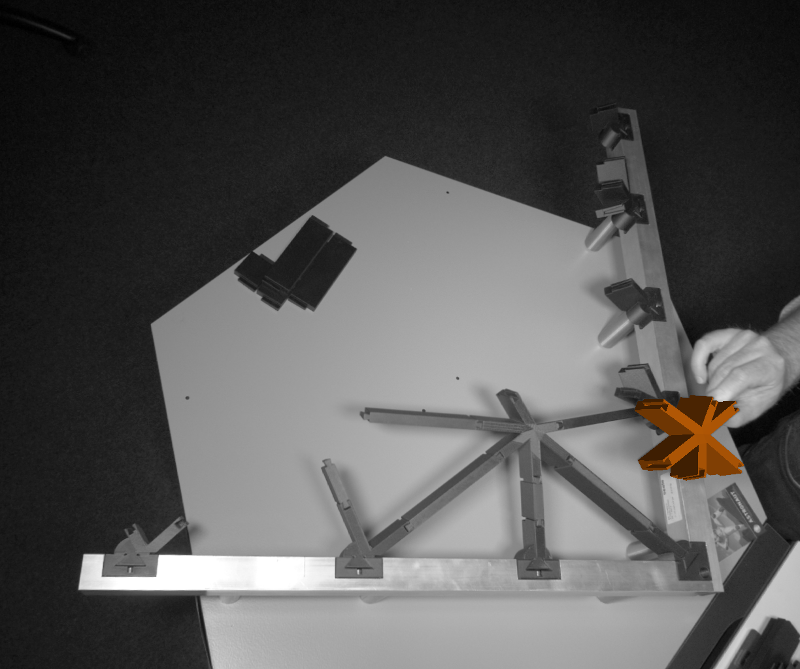}
\end{subfigure}
\hspace{.005\linewidth}
\begin{subfigure}{.23\linewidth }
  \centering
  \includegraphics[width=\linewidth]{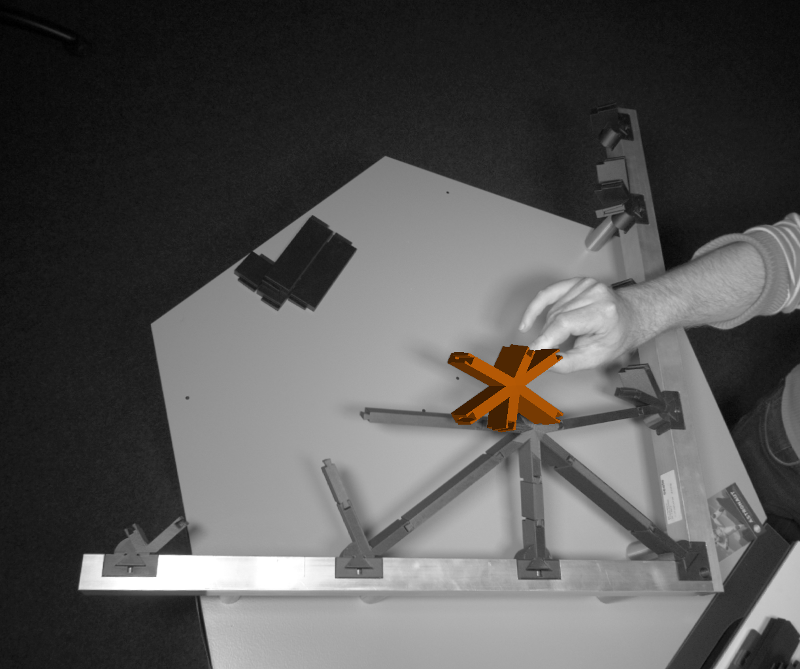}
\end{subfigure}
\hspace{.005\linewidth}
\begin{subfigure}{.23\linewidth }
  \centering
  \includegraphics[width=\linewidth]{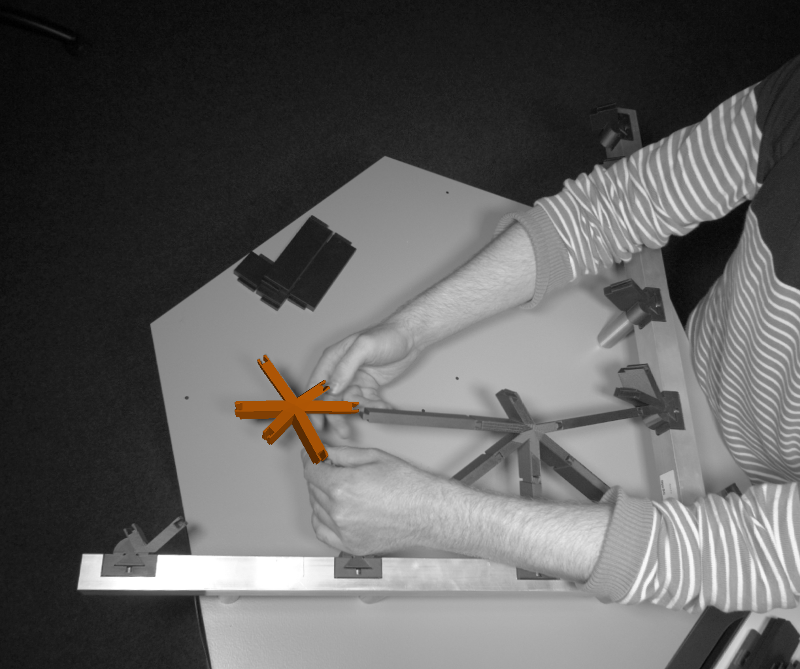}
\end{subfigure}
\centering  
\captionsetup{justification=centering}
\caption{Tracking an object in \textit{Far-Range} mode. The initialization area is highlighted by a green box.}
\label{fig:farrange}
\end{figure*}

\paragraph{Far-Range mode.}
We also use the approach to track objects from a bird's eye static camera that captures their assembly in a target structure (\autoref{fig:farrange}). Only one object is tracked at a time, so the detection network is not needed at all, as long as local tracking succeeds. In the images, the object appears much smaller than in training data. To compensate for this, the CNN is applied to the content of a square bounding box around the last detection or a predefined start position. In our setup, its side length is half the image size. Alternatively, it could be scaled dynamically based on the size of the object in the previous frame. An additional global detection network is avoided. Of course, it would also be possible to depict a larger variation of object-camera distances in the training dataset. The proposed reconfiguration should not be seen as a limitation, but as a way to use the trained network more flexibly. We found that for more distant objects only two instead of three pyramid levels and a lower value of $e_{init}$ are preferable. 

\subsection{Performance}
An inference of our CNN with OpenCV requires $\sim$20 ms on an Nvidia GTX 2080 Ti and image size $400^2$. When the network is used for each frame in \textit{Close-Range} mode, frame rates of 25-30 frames per second are achieved (slightly depending on the size of the object in the image) when a single object is also tracked locally.
We currently execute the local refinement sequentially, so the performance degrades if more objects are present. 
One way to improve that is to render an extra mask of object IDs and then do the optimization step for multiple objects simultaneously. Thereby, occlusions between known objects can be explicitly modelled and further improve the local tracking \cite{Huang2020}. High frame rates keep the distance between frames small and therefore stabilize local tracking, with less need for reinitialization. If, like in the \textit{Far-Range} mode, the network is not evaluated for every image and the object is small compared to the image size, the local refinement runs at $\sim$60 fps.

\section{\uppercase{Conclusion}}
We have presented a pipeline for multiple object detection and tracking that combines the advantages of current CNNs for 6D pose estimation with a local pose optimizer and a reliable metric for triggering reinitialization.
We use only synthetic training data, allowing fully automatic training without manual image labeling, and show that local-refinement is suitable to bridge the domain gap. In addition, we have presented an extension for a state-of-the-art CNN to better handle multiple, potentially similar objects, and demonstrate its benefits using a set of 13 similar, uncolored and non-textured objects. 

In real sequences, difficult situations not encountered in training can be better handled with a combined approach that draws on knowledge from previous images. We argue that such combined approaches are a very useful option for AR systems. The advantage of the easy generation of training data outweighs the extra effort of additional local optimization. The system is already being used in an AR demonstrator, although under relatively controlled conditions. In future work, we will further improve robustness by explicitly modelling occlusions between objects and possibly lift the self-imposed restriction of using only grayscale images.

\section*{\uppercase{Acknowledgments}}

This work is supported by the German Federal Ministry of Economic Affairs and Energy 
(DigitalTWIN, grant no. 01MD18008B and BIMKIT, grant no. 01MK21001H). We thank Fabian Schmid, Philipp Kopriwa, and Gergey Matl (se commerce GmbH) for the pleasant collaboration and for providing the reference models.

\bibliographystyle{apalike}
{\small
\bibliography{literature}}



\end{document}